\def\year{2020}\relax
\documentclass[letterpaper]{article} 
\usepackage{aaai20}  
\usepackage{times}  
\usepackage{helvet} 
\usepackage{courier}  
\usepackage[hyphens]{url}  
\usepackage{graphicx} 
\usepackage{graphicx}  
\title{Learning to Model Opponent Learning (Student Abstract)}
\author{\Large \textbf{Ian Davies\textsuperscript{\rm 1}, Zheng Tian, Jun Wang} \\ 
University College London\\ 
Gower Street\\
London, United Kingdom\\
WC1E 6BT\\
\textsuperscript{\rm 1}ian.davies.12@ucl.ac.uk 
}
\usepackage{subfigure}
\usepackage{booktabs}
\usepackage{microtype}

\newcommand{\centeredd}[1]{\begingroup\begin{tabular}{c} #1 \end{tabular}\endgroup}
\newcommand{\argmax}{\mathop{\mathrm{argmax}}}
\usepackage{amsfonts}
\usepackage{amsmath}
\usepackage{amssymb}
\usepackage{placeins}

\usepackage{multibib}
\newcites{main}{References}

\usepackage[bottom]{footmisc}
\begin{document}
\maketitle

\begin{abstract}
    Multi-Agent Reinforcement Learning (MARL) considers settings in which a set of coexisting agents interact with one another and their environment. The adaptation and learning of other agents induces non-stationarity in the environment dynamics. This poses a great challenge for value function-based algorithms whose convergence usually relies on the assumption of a stationary environment. Policy search algorithms also struggle in multi-agent settings as the partial observability resulting from an opponent's actions not being known introduces high variance to policy training. Modelling an agent's opponent(s) is often pursued as a means of resolving the issues arising from the coexistence of learning opponents. An opponent model provides an agent with some ability to reason about other agents to aid its own decision making. Most prior works learn an opponent model by assuming the opponent is employing a stationary policy or switching between a set of stationary policies. Such an approach can reduce the variance of training signals for policy search algorithms. However, in the multi-agent setting, agents have an incentive to continually adapt and learn. This means that the assumptions concerning opponent stationarity are unrealistic. In this work, we develop a novel approach to modelling an opponent's learning dynamics which we term Learning to Model Opponent Learning (LeMOL). We show our structured opponent model is more accurate and stable than naive behaviour cloning baselines. We further show that opponent modelling can improve the performance of algorithmic agents in multi-agent settings.
\end{abstract}

In the context of multi-agent reinforcement learning, modelling an opponent can take many forms including inferring an opponent's motivation, representing an opponent through underlying characteristics and learning to predict an opponent's actions. Our work is concerned with action prediction, drawing from works on agent representation and meta-learning to model an agent's evolution throughout learning.

Previous works have considered adapting to a non-stationary opponent by learning a new policy once the opponent is perceived to have changed \citemain{zheng2018deep}. Such a setting requires the opponent to play a stationary policy while an effective response is learned. These prior approaches to playing with a non-stationary opponent do not consider the structure of the non-stationarity of an opponent. Our work aims to exploit the structure of an opponent's learning process to continuously adapt to a learning opponent. This is a fundamental and challenging issue in multi-agent reinforcement learning \citemain{hernandez2017survey}.

Conditioning an agent's policy upon the (predicted) action of an opponent stabilises policy updates. This follows from the update being specific to the gradient of the loss at a particular observation-opponent action pair. Accounting for the opponent's action means that, for different opponent actions the policy acts and is updated accordingly.

In the decentralised setting, where the opponent's policy cannot be freely accessed to attain true actions, a sufficiently performant opponent model has the potential to overcome the loss of information from decentralisation and therefore enable decentralised training. Decentralisation through action prediction would be a key advancement in multi-agent reinforcement learning.

\begin{figure*}[t!]
  \centering
  \subfigure[Prediction Cross-Entropy During Play]{\centering\includegraphics[width=0.32\textwidth]{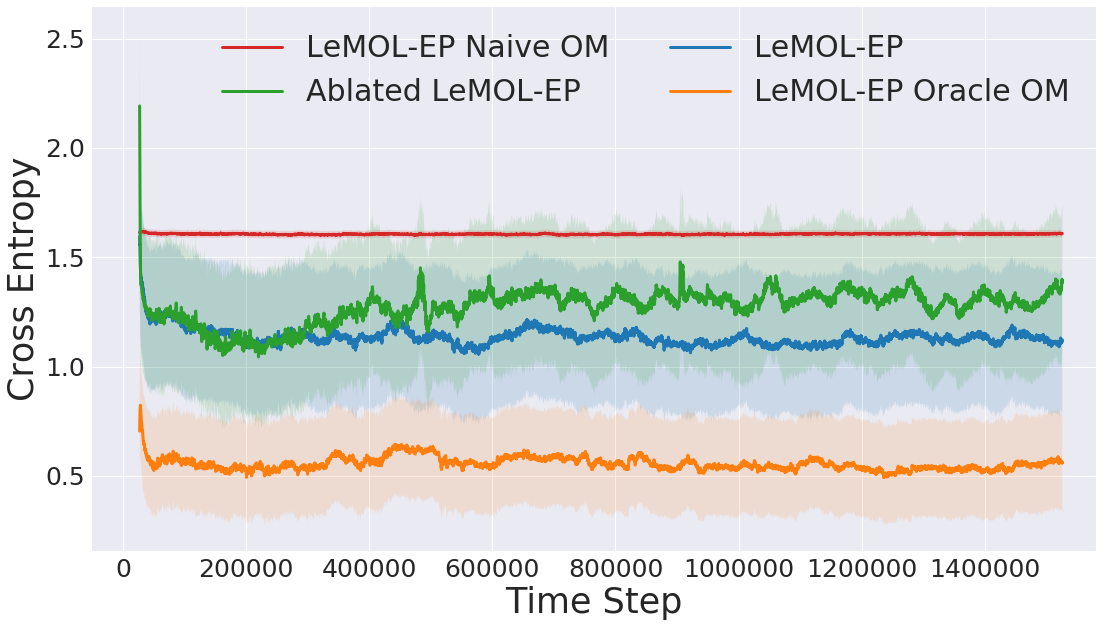}\label{fig:lemol-xent}}
  \subfigure[Centralised LeMOL-EP Results]{\centering\includegraphics[width=0.32\textwidth]{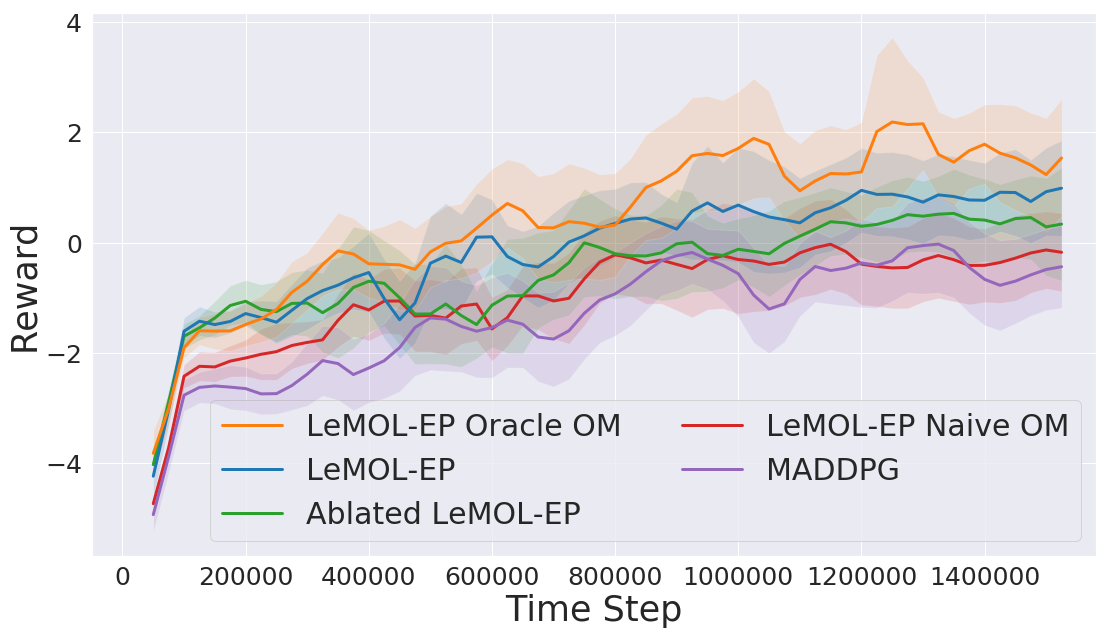}\label{fig:lemol-ep-rew}}
  \subfigure[Decentralised LeMOL-EP Results]{\centering\includegraphics[width=0.32\textwidth]{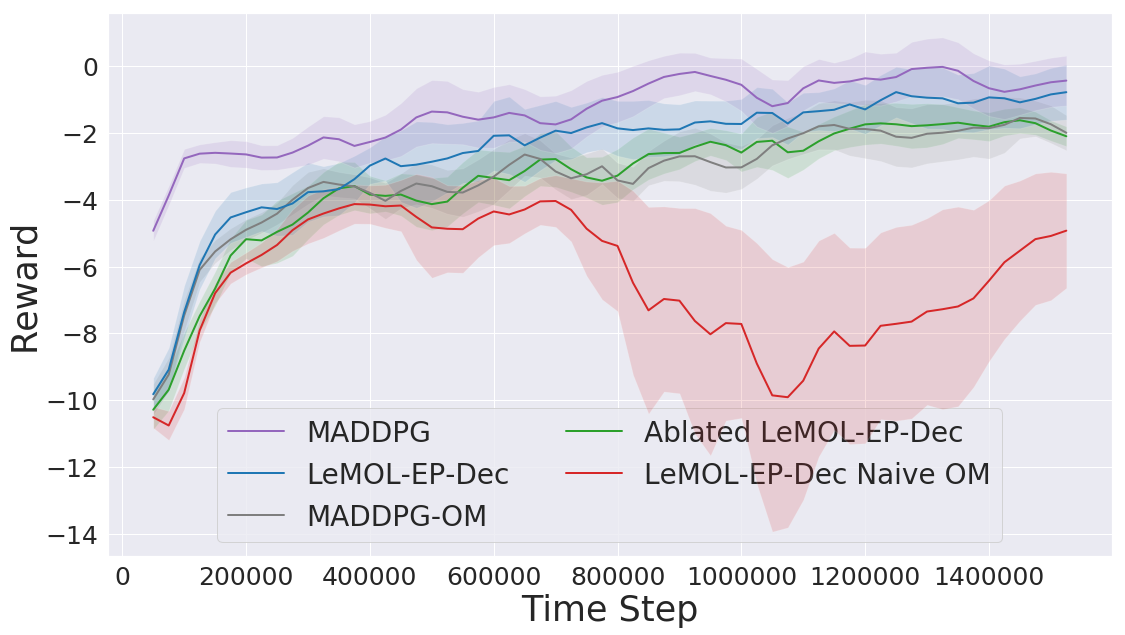}\label{fig:lemol-ep-dec-rew}}
  \caption{Results from experiments with LeMOL-EP in the centralised and decentralised setting. Solid lines are averages (mean) from 15 runs. Shaded regions denote one standard deviation.}
  \label{fig:summary-of-results}
\end{figure*}

\section{Methodology}

We augment the centralised actor-critic architecture of multi-agent deep deterministic policy gradients (MADDPG) \citemain{lowe2017maddpg} with a novel opponent model based on the meta-learning algorithm RL$^2$ \citemain{duan2016rl2}. RL$^2$ is based on an LSTM network which stores the state of a task-specific agent in its activations. The role of the LSTM is to adapt the task-specific agent to a new task. The LSTM is trained to learn a generalisable update rule for its hidden state which can replace closed-form gradient descent techniques for training on new tasks. The state update rule of the LSTM therefore becomes an optimisation algorithm trained on the performance of the agents it generates in varied environments.

We aim to emulate an opponent's learning rather than learn a generalisable optimisation technique. In light of this, our recurrent module stores and updates a representation of the opponent. The state update function is therefore trained to emulate the opponent's learning. This training is treated as a regression problem for predicting opponent actions from the observed history of the game. We utilise a method we term Episode Processing (EP) whereby each episode of experience is summarised by a bidirectional LSTM and is then used to update our agent's representation of its opponent.

\section{Experiments}
We use the Open AI particle environments \citemain{lowe2017maddpg} for experiments. Specifically, we focus on the two-player adversarial game Keep-Away.

Our LeMOL agents are endowed with an in-episode LSTM network to aid with the issue of partial observability. Our agents take on the role of the defender trying to keep the attacker away from the goal. The goal is one of two landmarks and the defender does not know which. The attacker is trained using MADDPG.

Our experiments compare our MADDPG baseline, our full LeMOL-EP model, LeMOL-EP where modelling of the opponent's learning process is removed (Ablated LeMOL-EP), LeMOL-EP where the opponent model has perfect prediction accuracy (LeMOL-EP Oracle OM) and LeMOL-EP where the opponent model is untrained (LeMOL-EP Naive OM). In the decentralised setting we also include a version of MADDPG with opponent modelling to make it amenable to the decentralised setting (MADDPG-OM).

\section{Results}
Comparison of opponent model performance for the full and ablated LeMOL-EP models in Figure \ref{fig:lemol-xent} demonstrates the benefit, in terms of action prediction accuracy, of modelling the opponent's learning. Having a continuously updated opponent model improves and stabilises opponent model performance. Figure \ref{fig:lemol-ep-rew} shows the impact of improved opponent modelling on agent performance. We find that the reduction in the variance of policy updates resulting from conditioning an agent's policy on predictions of the opponent's actions improves overall agent performance.

In the decentralised setting (Figure \ref{fig:lemol-ep-dec-rew}), we find that using the opponent model can enable effective decentralised training, as the opponent model compensates for the inability to access to others' policies in the decentralised setting. Note that the architecture of the opponent models is consistent across centralised and decentralised settings. Our decentralised model attains a similar level of performance to the centralised MADDPG agent. The opponent model is the only means of accounting for non-stationarity under decentralised training. Results are therefore highly sensitive to the accuracy of the opponent model. This is demonstrated by the instability and poor performance of the model with a naive (untrained) opponent model. This model collapses back to a single agent approach ignoring the opponent's presence.

We find that the more accurate an opponent model, the greater the improvement in agent performance. This is particularly pronounced in the decentralised setting where the increased opponent model accuracy and stability provided by modelling the opponent's learning process is essential to attain similar performance to centralised MADDPG.

\section{Directions for Future Work}
This work provides initial evidence for the efficacy of modelling opponent learning as a solution to the issue of non-stationarity in multi-agent systems. Furthermore, we have shown that such modelling improves agent performance over the strong MADDPG baseline in the centralised setting. When our approach is applied to decentralised training it achieves comparable performance to the popular centralised MADDPG algorithm.

Despite these promising results there is significant work to be done to extend and enhance the framework we develop for handling non-stationarity through opponent modelling. We hope to pursue a formal Bayesian approach to opponent learning process modelling in future. We hope such an approach will enable a theoretical framework to emerge which can be validated through further experiments.

\bibliographystylemain{aaai}
\bibliographymain{References}

\onecolumn
\appendix

\setcounter{secnumdepth}{2}
\section{Methodology Details \& Model Architecture\footnote{Code for the implementation of LeMOL and baselines is available at \texttt{https://github.com/ianRDavies/LeMOL}}}\label{appendix:architecture}

\begin{figure*}[b!]
  \centering
  \subfigure[Episode Processing]{\centering\includegraphics[width=0.75\textwidth]{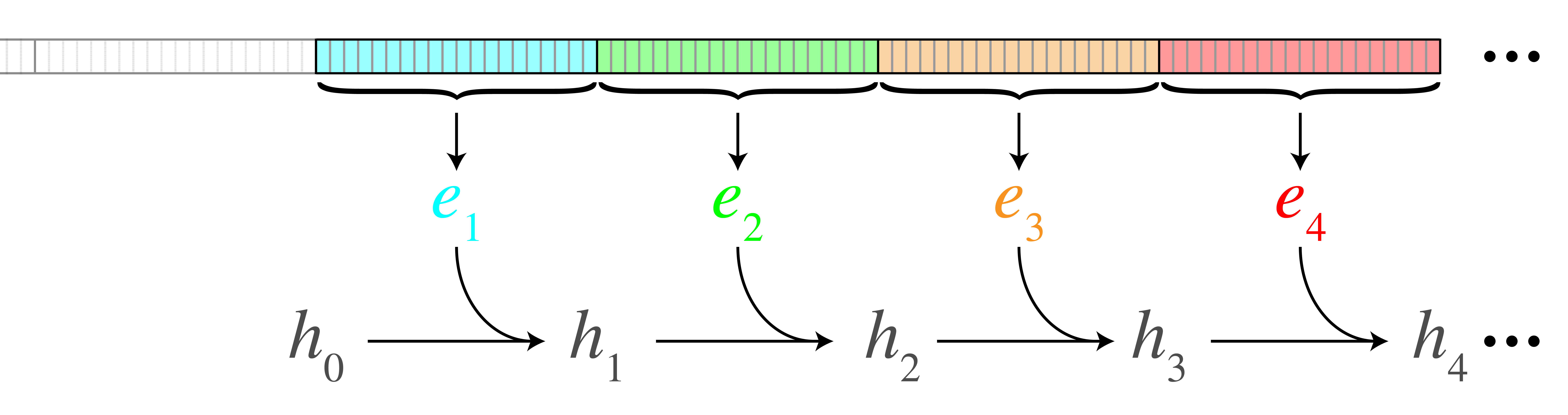}\label{fig:episode-processing}}
  \subfigure[Opponent Model Structure]{\centering\includegraphics[width=0.45\textwidth]{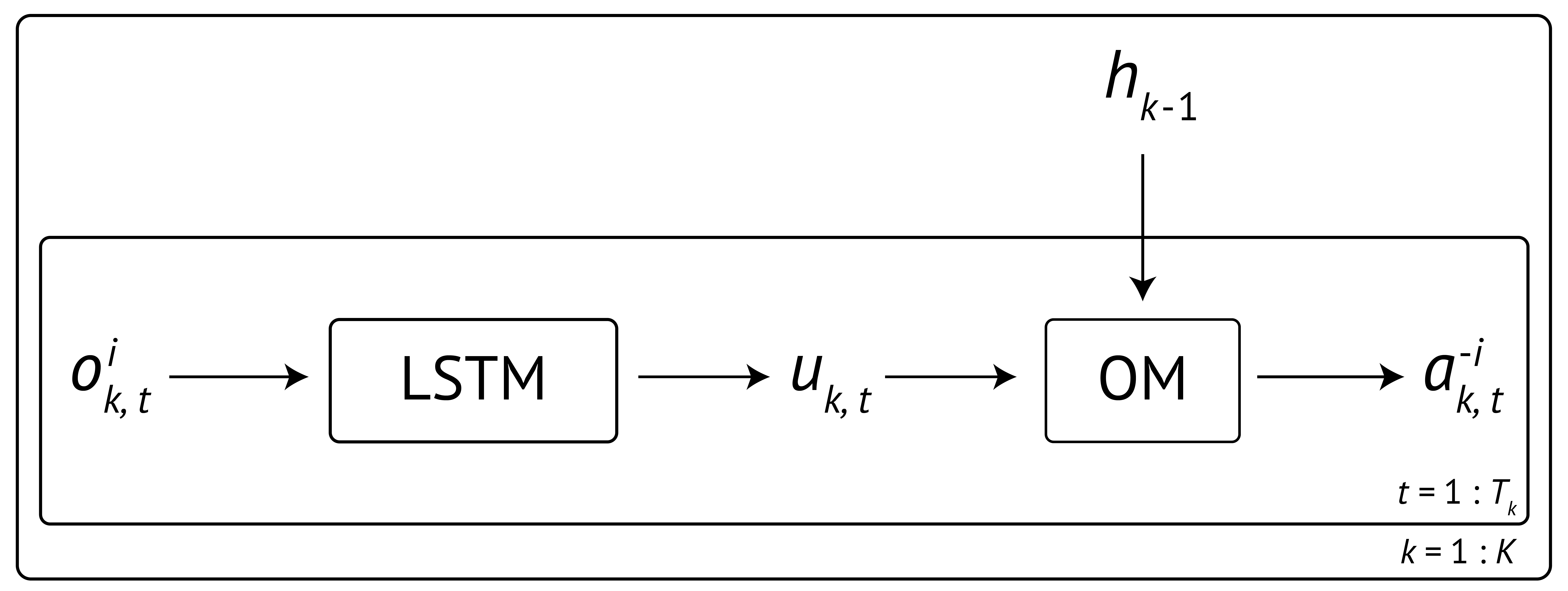}\label{fig:in-ep-lstm}}
  \caption{Learning to Model Opponent Learning with Episode Processing (LeMOL-EP) architecture. a) The episodes after initial exploration (each with their own colour) are summarised by a bidirectional LSTM to produce episode summaries $e_k$. The summary for the $k^\text{th}$ episode is used to update the opponent representation $h_k$. b) The opponent model itself is formed from an LSTM which processes observations to produce an in-episode history summary $u_{k,t}$. The prediction function then takes a summary of the episode history $u_{k,t}$ and the opponent representation $h_{k-1}$ as inputs and predicts the opponent's action $a^{-i}_{k, t}$. $k$ indexes episodes and $t$ indexes time steps within each episode.}
  \label{fig:architecture-diagrams}
\end{figure*}
Our proposed architecture utilises the structure of the opponent's learning process to develop a continuously adapting opponent model. We build a model which, by using an opponent representation updated in line with the opponent's learning, achieves truly continuous adaptation.

We focus on the competitive setting where agents' objectives are opposed and information is kept private. We therefore build an opponent model architecture capable of learning to account for the influence of the opponent's learning on its behaviour from the observed history of the game alone.

\subsection{Opponent Modelling}
We pursue explicit opponent modelling, where the opponent model is used to predict the action of the opponent at each time step. The opponent's policy is modelled as a distribution over actions.

We handle the non-stationarity of the opponent's policy and the in-episode partial observability separately in what we term \emph{Episode Processing (EP)}. This approach makes modelling of the opponent's learning process explicit and conditions the prediction of the opponent's action on the outcome of the model of the opponent's learning process.

We assume that opponents update periodically, once per episode. In Figure \ref{fig:episode-processing} each episode is highlighted in a different colour. Within each episode, agents' policies are not updated and therefore the opponent's policy is stationary such that opponent actions vary only due to the current state of play and policy stochasticity. The learning of the opponent then has an impact only in inter-episode variation. In order to capture this in our model of opponent learning, we first form a representation of the events of each episode, denoted by $e_k$ for the $k^\text{th}$ episode. These episode summaries then form a time series $e_{1:K}$ which traces the opponent's learning trajectory. This time series is the input to an LSTM which updates its state to track the opponent's evolution through learning.

During training, a LeMOL agent plays a full set of episodes, which we term a (learning) trajectory, such that both the LeMOL agent and its opponent train their policies and critics to convergence. After data from several trajectories, $\{\tau_m\}_{m=0}^M$ have been collected, LeMOL's opponent model is trained (via backpropagation) to model opponent learning and predict opponent actions using the previously experienced trajectories as training data.

We propose the use of a recurrent neural network (RNN) to model an opponent's learning process. By learning an update rule for the state of the RNN, $h$, our agents are able to process and preserve experience across a full opponent learning trajectory. Specifically we choose a Long Short-Term Memory (LSTM) \cite{hochreiter1997lstm} architecture for our RNN. This approach is inspired by the $\text{RL}^2$ algorithm \cite{duan2016rl2} which uses an LSTM to track and update a policy for a reinforcement learning agent. For our opponent model, the hidden activations of the LSTM, $h$, form a representation of the opponent and the learned state update rule of the LSTM aims to emulate the opponent's learning rule.

The episode summaries, $e_{1:K}$, used for modelling the opponent's learning process are produced by a bidirectional LSTM network \cite{schuster1997bidirectional,graves2005framewise}. The input to the bidirectional LSTM is the full set of events of the episode in question. We define an event at time step $t$ of episode $k$, to be the concatenation of an agent's observation $o^i_{k, t}$, action $a^i_{k, t}$ and reward $r^i_{k, t}$ with the observed opponent action $a^{-i}_{k, t}$ and shared termination signal $d_{k, t}$. We denote events by $\mathbf{x}^k_t$. We use $a:b$ to denote the set of episodes or time steps from the $a^\text{th}$ to the $b^\text{th}$ inclusive of the end points $a$ and $b$. 

The process for modelling the opponent's learning process through episode summaries can be considered as two subprocesses: i) summarise the episode within which policies are fixed using a bidirectional LSTM (Equation \ref{eq:lemol-ep-summary}), ii) use the episode summary $e_k$ to update the hidden state of the opponent model's core LSTM which tracks opponent learning (Equation \ref{eq:lemol-block-h-update}). These processes are depicted in Figure \ref{fig:architecture-diagrams}.
\begin{align}
    e_k &= \text{BidirectionalLSTM}(\mathbf{x}^k_{1:T})\label{eq:lemol-ep-summary}\\
    h_k &= f_{\psi_\text{LSTM}}\left(e_k, h_{k-1}\right) \label{eq:lemol-block-h-update}
\end{align}
where $\mathbf{x}_{1:T_k}$ denotes the sequence of events that make up the $k^\text{th}$ episode.

To handle partial observability from the environment, we introduce an in-episode LSTM which processes the incoming stream of observations during play to maintain an in-episode observation history summary $u_{k,t}$.

The in-episode LSTM is reset at the start of each episode such that the memory of the action prediction function is limited to the current episode. The operation of the in-episode LSTM is summarised by Equations \ref{eq:in-ep-lstm} and \ref{eq:recurrent-om} and is depicted by Figure \ref{fig:in-ep-lstm}.
\begin{align}
    u_{k,t} &= f_{\psi_\text{in-ep}}\left(o^i_{k,t}, u_{k,t-1}\right) \label{eq:in-ep-lstm}\\
    \hat{a}_{k, t}^{-i} &= \argmax_{a^{-i}} \rho_{ \psi_\text{OM} }\left(a^{-i}\middle|u_{k,t}, h_{k-1}\right)
    \label{eq:recurrent-om}
\end{align}

We train the opponent model using the chosen action target (CAT) formed from the opponents' actions. This choice of target yields the loss function in Equation \ref{eq:l-cat} which is amenable to decentralised opponent model training, under the assumption that the opponent's actions are observed as they are executed. This objective is one of maximum likelihood.
\begin{align}
    \mathcal { L } _ { \mathrm { CAT } }(\psi_\text{OM}^i) = - \sum_{t} \mathbb{E}_{o^i_{k,t}, \mathbf{x}^{1:{k-1}}_{1:T}}\left[\log \left( \rho_{\psi_\text{OM}}\left(a^{-i}_{k,t}\middle|o_{k,t}^i, h_{k-1}\left(\mathbf{x}^{1:{k-1}}_{1:T}\right)\right) \right)\right]\label{eq:l-cat}
\end{align}

In practice, the integral required to calculate the expectation in Equation \ref{eq:l-cat} is intractable due to the possibly high dimension of the observation space. We therefore use Monte Carlo sampling to replace the integral in the expectation leading to the tractable loss function in Equation \ref{eq:l-cat-tract}.

\begin{align}
    \mathcal{L}_{\text{CAT}}(\psi_\text{OM}) &= -\sum_{t}\mathbb{E}_{o^i_{k,t}, \mathbf{x}^{1:{k-1}}_{1:T}}\left[\log\left[ \rho_{\psi_\text{OM}}\left(a^{-i}_{k,t}|o_{k,t}^i, h_{k-1}\right) \right]\right] \nonumber\\
    &=-\sum_t\int\int\log\left( \rho_{\psi_\text{OM}}\left(a^{-i}_{k,t}\middle|o_{k,t}^i, h_{k-1}\right) \right)\ d\mathbf{x}^{1:k-1}_{1:T}\ d o_{k,t}^i\nonumber\\
    &\approx -\frac{1}{M}\sum_m\sum_t\log\left( \rho_{\psi_\text{OM}}\left.\left(a^{-i}_{t,m}\right|o_{t,m}^{i}, h_{t,m}\right) \right)\label{eq:l-cat-tract}
\end{align}

This requires sampling a minibatch of $M$ full learning trajectories which are processed in sequence so that the LSTM can effecively maintain its hidden state.

Our opponent model is part of an actor-critic approach to training. We incorporate our novel opponent model into the MADDPG actor-critic framework proposed by \citeauthor{lowe2017maddpg}~\shortcite{lowe2017maddpg}.

\subsection{The Critic}
We use a centralised Q function that takes inputs from all agents to stabilise training and improve performance.\footnote{In this work we use $s_+$ and $a_+$ rather than $s'$ and $a'$ to denote subsequent states and actions in order to tidy up notation.}
\begin{equation*}
    Q_{\theta^i}\left(o^i, o^{-i}, a^i, a^{-i}\right) = r^i_s\left(a^i, a^{-i}\right) + \gamma\mathbb{E}_{s_+(a^i, a^{-i})}\left[V\left(s_+\right)\middle|a^i\sim\pi^i, a^{-i}\sim\pi^{-i}\right]
\end{equation*}

Our Q function is parameterised as a modestly sized neural network which outputs a single value given the observations of all agents as well as their actions. $V(s)$ denotes the value of a state.

We train the Q function to minimise the temporal difference error defined in 
Equation \ref{eq:lemol-q-loss}.
\begin{equation}
    \mathcal{L}_Q(\theta^i) = \left(
    Q_{\theta^i}\left(o^i, o^{-i}, a^i, a^{-i}\right) - y
    \right)^2\hspace{0.5cm}\text{where}\hspace{0.5cm}
    y = r^i + \gamma \bar{Q}_{\bar{\theta}^i} \left(o^i_+, o^{-i}_+, \tilde{a}^i_+, \tilde{a}^{-i}_+\right)\label{eq:lemol-q-loss}
\end{equation}
where $r^i$ denotes the reward received by agent $i$ and $o^i_+, a^{i}_+, o^{-i}_+$ and $a^{-i}_+$ denote the observations and actions at the next time step for agents $i$ and $-i$ respectively. Furthermore, $\bar{Q}_{\bar{\theta}}$ denotes a target network and $\tilde{a}^i_+$ and $\tilde{a}^{-i}_+$ are actions (given the next observations) output by target policy networks.

The target networks are independently initialised from the main networks of the model and are slowly updated by Polyak averaging \cite{polyak1990new}, this is an approach to stabilise training by avoiding the optimism induced by bootstrapping from a Q function which is itself the subject of training.

\subsection{Policy Training}
As is conventional in actor-critic settings, we use a policy gradient approach to learn an effective policy. This leads us to use the Q function to form an objective for the policy to minimise by following the gradient with respect to the policy parameters, $\phi^i$. The objective in the case of LeMOL is given in Equation \ref{eq:lemol-pi-loss}, where the expectation is taken over the actions from the policy being trained with the opponent's actions and the observations being drawn jointly from a replay buffer $\mathcal{D}$ which stores the observed history of the game.
\begin{equation}
    \mathcal{L}_{\pi}(\phi^i)= - \mathbb{E}_{a^i\sim\pi^i_{\phi^i}, (o^i, o^{-i}, a^{-i})\sim\mathcal{D}}\left[Q_{\theta^i}\left(o^i, o^{-i}, a^i, a^{-i}\right)\right]
    \label{eq:lemol-pi-loss}
\end{equation}



\subsection{Decentralised Training Through Action Prediction}

In the decentralised setting, we are no longer able to utilise the opponent's observation as an input into the Q function. The Q function for decentralised training is therefore a function of the actions of both agents and the observation of the LeMOL agent only.
\begin{equation*}
Q_{\theta^i}\left(o^i, a^i, a^{-i}\right)=r^i_s\left(a^i, a^{-i}\right) + \gamma\mathbb{E}_{s_+(a^i, a^{-i})}\left[V\left(s_+\right)\middle|a^i\sim\pi^i, a^{-i}\sim\pi^{-i}\right]
\end{equation*}
During training of the Q function, previous observations, actions, opponent actions and rewards are sampled from the replay buffer. The Q function target, denoted by $y$ (Equation \ref{eq:decentalised-lemol-q-loss}), is then calculated using the sampled reward and an evaluation of the target Q-function.
\begin{gather}
    \hat{a}^{-i}_+ = \argmax_{a^{-i}}\rho_{\psi_\text{OM}^i}\left(a^{-i}\middle|u_+, h_+\right)\\
    \tilde{a}^i_+ \sim \bar{\pi}_{\bar{\phi}^i}\left(o^i, \hat{a}^{-i}_+\right)\\
    \mathcal{L}_Q(\theta^i) = \left(
    Q_{\theta^i}\left(o^i, a^i, a^{-i}\right) - y
    \right)^2\hspace{0.7cm}\text{where}\hspace{0.7cm}
    y = r^i + \gamma \bar{Q}_{\bar{\theta}^i} \left(o^i_+, \tilde{a}^i_+, \hat{a}^{-i}_+\right)\label{eq:decentalised-lemol-q-loss}
\end{gather}

Note that the opponent model is passed the latest representation of the opponent, $h_k$, as an input. Decentralisation means that it is no longer possible to query the opponent's target policy network. Therefore we must use the opponent model to generate an opponent action to input. Utilising this predicted action for the opponent in the target Q function evaluation enables the Q function target to reflect the opponent's play as it would be given its present state of learning. Our opponent model makes this possible without relying on direct access to the opponent's (target) policy.

In the decentralised setting, policy training uses the decentralised Q function as a critic. We predict the opponent's action for the state as observed by our agent and use this in place of the historical opponent action sampled from the replay buffer. This means that the policy is trained using the Q function evaluated at the action profile of the agents at their present state of learning. In this way, an accurate opponent model enables us to train our agent using the game history reevaluated for the updated opponent.
\begin{gather*}
    \mathcal{L}_\pi(\phi^i)= - \mathbb{E}_{a^i\sim\pi^i_{\phi^i}, (o^i, u, h)\sim\mathcal{D}}\left[Q_\theta\left(o^i, a^i, \hat{a}^{-i}\right)\right]\hspace{0.5cm} \text{where}\hspace{0.5cm}{\hat{a}^{-i}=\argmax_{a^{-i}}\rho_{\psi^i_\text{OM}}\left(a^{-i}\middle|u, h\right)}\\\small
    \nabla_{\phi^i} J(\phi^i)=\mathbb{E}_{(o^i, u, h)\sim\mathcal{D}}\left[\nabla_{\phi^i}\pi^i_{\phi^i}\left(a^i\middle|a^i,\hat{a}^{-i}\right)\nabla_{a^i}Q_\theta\left(o^i, a^i, \hat{a}^{-i}\right)\right]
\end{gather*}
where
\begin{equation*}
    {a^i\sim\pi^i_{\phi^i}(o^i, \hat{a}^{-i})\hspace{0.5cm}\text{and}\hspace{0.5cm} \hat{a}^{-i}=\argmax_{a^{-i}}\rho_{\psi_\text{OM}}(a^{-i}|u, h)}
\end{equation*}

\clearpage

\section{Related Work}\label{appendix:related-work}
Previous works seeking to handle the non-stationarity of opponents in multi-agent reinforcement learning have sought to relearn policies once the opponent is perceived to have changed \cite{zheng2018deep,everett2018learning}. Such approaches rely on the opponent playing a stationary policy while they are being modelled so that an effective model for play and change detection can be trained. This approach is not suited to a continually learning opponent as present in our work. We utilise the structure of an opponent's learning process to learn a series of opponent representations which can be used to inform opponent action prediction. We therefore avoid having to relearn a policy as we learn a single policy which handles non-stationarity by taking in a prediction of the opponent's action.

\citeauthor{al2017continuous}~\shortcite{al2017continuous} take an alternative meta-learning inspired approach to effectively learn many models at once. Their agents adapt to differing opponents which are stationary within a given period. Their approach is an application of model-agnostic meta-learning \cite{finn2017model} such that a base set of parameters are learned and adapted anew for different stages of an opponent's training. Their agents therefore require an additional set of optimisation steps for each new period of play. Their approach also does not take advantage of the structured and sequential nature of an opponent's learning.

\citeauthor{hong2018drpiqn}~\shortcite{hong2018drpiqn} present two algorithms Deep Policy Inference Q-Network (DPIQN) and a recurrent version (DRPIQN). Their approach updates Q learning with an auxiliary implicit opponent modelling objective. However, they do not explicitly consider the non-stationarity of the opponent. Their experiments have deterministic opponents which switch between different policies during play. They also consider the case of a learning teammate but do not consider the learning of an adversarial opponent. In future work, we would hope to include DRPIQN as a comparison benchmark to see how it performs in the competitive setting with a learning opponent. Such experiments would echo our MADDPG-OM and decentralised LeMOL-EP experiments which contrast opponent modelling for Q-learning alone (as in D(R)PIQN and MADDPG-OM) and opponent modelling for both policy and Q-learning (as for LeMOL-EP). Comparison to D(R)PIQN would enable us to consider whether predicting opponent actions is the best way to use an opponent model. The performance of agents with implicit opponent models could benefit from capturing opponent characteristics beyond their current action selection.

Both model switching and meta-learning works have been restricted to considering opponents which change in distinct discrete steps, between which the opponent is stationary. By doing so, previous works have been able to develop theoretical grounding for their models and achieve experimental success. We appeal to the structure of an opponent's learning process and draw on the meta-learning literature to develop truly continuous adaptation. We believe that this is a novel and challenging line of work which warrants more attention.

One related work which considers the structure of an opponent's learning architecture is Learning with Opponent Learning Awareness (LOLA) \cite{foerster2018learning}. LOLA agents account for the impact of their own policy updates on their opponent's policy by differentiating through their opponent's policy updates. The success of LOLA agents is dependent on self play of two homogeneous agents so that the impact of one agent's policy on the learning step of the other agent(s) can be calculated or estimated. Our approach is agnostic to the architecture and learning methodology of the opponent.

\section{Experiments}\label{appendix:experiments}
The features of the models used in our experiments are laid out in Table \ref{tbl:model-features}. Each model is run for 15 full trajectories against an MADDPG opponent. The hyperparameters used to train our models are kept fixed and laid out in Tables \ref{tbl:hyperparams} and \ref{tbl:om-hyperparams}. Note that the relevant parameters are the same for MADDPG when used as a defender as well as an attacker (the opponent).

\vspace{-0.2cm}
\begingroup
\begin{table}[hb]
    \centering
    \begin{tabular}{ccccccc}
    \toprule
         \small &  \centeredd{\small On-Policy} & \centeredd{\small Centralised} & \centeredd{OM} & \centeredd{\small $\hat{a}^{-i}\to\pi^i$} & \centeredd{\small In-Episode\\\small OM LSTM} & \centeredd{\small Model of Opponent\\\small Learning Process}\\
    \midrule
         \small MADDPG & $\boldsymbol{\times}$ & $\checkmark$ & $\boldsymbol{\times}$ &  $\boldsymbol{\times}$ & $\boldsymbol{\times}$ & $\boldsymbol{\times}$\\
         \small MADDPG-OM & $\boldsymbol{\times}$ & $\boldsymbol{\times}$ & $\checkmark$ &  $\boldsymbol{\times}$ & $\checkmark$ & $\boldsymbol{\times}$ \\
         \small LeMOL-EP & $\boldsymbol{\times}$ & $\checkmark$ & $\checkmark$ & $\checkmark$ & $\checkmark$ & $\checkmark$\\
         \small Ablated LeMOL-EP & $\boldsymbol{\times}$ & $\checkmark$ & $\checkmark$ & $\checkmark$ & $\checkmark$ & $\boldsymbol{\times}$\\
         \small LeMOL-EP Oracle OM & $\boldsymbol{\times}$ & $\checkmark$ & $\checkmark$ & $\checkmark$ & \textbf{--} & \textbf{--} \\
         \small LeMOL-EP-Dec  & $\boldsymbol{\times}$ & $\boldsymbol{\times}$ & $\checkmark$ & $\checkmark$ & $\checkmark$ & $\checkmark$\\
         Ablated LeMOL-EP-Dec & $\boldsymbol{\times}$ & $\boldsymbol{\times}$ & $\checkmark$ & $\checkmark$ & $\checkmark$ & $\boldsymbol{\times}$ \\
    \bottomrule
    \end{tabular}
    \caption{The features of the models included in our experiments. OM stands for Opponent Model. The Dec suffix is used to denote decentralised implementations. $\hat{a}^{-i}\to\pi^i$ denotes the use of opponent action predictions as inputs to the agent's policy. The symbol \textbf{--} is used where features are not applicable due to the oracle having no defined structure as it simply passes values from the opponent to LeMOL-EP. Naive (untrained) LeMOL models are not included as they are equivalent in architecture to the full LeMOL models. They differ only in that the opponent model is not trained.}
    \label{tbl:model-features}
\end{table}
\endgroup

\begingroup
\begin{table}[htb]
    \centering
    \begin{tabular}{cc}
    \toprule
    Hyperparameter & Value \\
    \midrule
    Episode Length & 25 \\
    Number of Episodes & 61024 \\
    Episodes of Exploration & 1024 \\
    Batch Size for Q Network and Policy Training & 1024 \\
    Policy Network Size & (64, 64, 5) \\
    Policy and Q Network Optimiser & ADAM \cite{kingma2014adam}\\
    Policy and Q Network Learning Rate & 0.01 \\
    Policy and Q Network ADAM Parameters $(\beta_1, \beta_2, \epsilon)$ & $(0.9, 0.999, 10^{-8})$ \\
    Q Network Size & (64, 64, 1) \\
    Polyak & 0.01 \\
    Policy Update Frequency (Time Steps) & 25\\
    Replay Buffer Capacity & 1,000,000\\
    \bottomrule
    \end{tabular}
    \vspace{-0.1cm}
    \caption{Hyperparameters for LeMOL-EP Experiments}
    \label{tbl:hyperparams}
\end{table}

\begin{table}[htb]
    \centering
    \begin{tabular}{cc}
    \toprule
    Hyperparameter & Value \\
    \midrule
    Opponent Model LSTM State Dimension & 64 \\
    Chunk Length for Opponent Model Training (Time Steps) & 500 \\
    Opponent Model Training Batch Size & 8 \\
    Opponent Model Training Epochs & 50 \\
    Episode Embedding Dimension & 128 \\
    In-Episode LSTM State Dimension & 32 \\
    Opponent Model Optimiser & ADAM \cite{kingma2014adam}\\
    Opponent Model Learning Rate & 0.001 \\
    Opponent Model ADAM Parameters $(\beta_1, \beta_2, \epsilon)$ & $(0.9, 0.999, 10^{-8})$\\
    \bottomrule
    \end{tabular}
    \vspace{-0.1cm}
    \caption{Hyperparameters for LeMOL-EP Opponent Model}
    \label{tbl:om-hyperparams}
\end{table}
\endgroup
\FloatBarrier
\bibliographystyle{aaai}
\vspace*{-1cm}
\bibliography{References}

\end{document}